\documentclass[conference]{IEEEtran}
\IEEEoverridecommandlockouts
\usepackage{cite}
\usepackage{amsmath,amssymb,amsfonts}
\usepackage{algorithmic}
\usepackage{graphicx}
\usepackage{textcomp}
\usepackage{xcolor}
\usepackage{mathtools}
\usepackage{booktabs}
\usepackage{tabularx}
\usepackage{array}
\usepackage{multirow}
\usepackage{float}
\usepackage{listings}
\usepackage{silence}
\usepackage{caption}
\usepackage[hidelinks]{hyperref}
\usepackage{microtype}
\newcolumntype{L}{>{\raggedright\arraybackslash}X}
\newcolumntype{P}[1]{>{\raggedright\arraybackslash}p{#1}}
\newcolumntype{B}[1]{>{\bfseries\raggedright\arraybackslash}p{#1}}
\usepackage[table]{xcolor}
\definecolor{siemensteal}{HTML}{009999}

\def\BibTeX{{\rm B\kern-.05em{\sc i\kern-.025em b}\kern-.08em
    T\kern-.1667em\lower.7ex\hbox{E}\kern-.125emX}}

\lstdefinelanguage{Cypher}{
  morekeywords={MATCH,MERGE,CREATE,UNWIND,RETURN,WITH,WHERE,ORDER,BY,ASC,
    DESC,OPTIONAL,FOREACH,CASE,WHEN,THEN,ELSE,END,AS,IN,IS,NOT,NULL,ON},
  sensitive=true,
  morecomment=[l]{//},
  morestring=[b]"
}
\lstdefinestyle{code}{
  basicstyle=\ttfamily\scriptsize,
  breaklines=true,
  frame=single,
  aboveskip=\medskipamount,
  columns=fullflexible,
  keepspaces=true,
  showstringspaces=false,
  backgroundcolor=\color{gray!8},
  keywordstyle=\color{blue!70!black}\bfseries,
  commentstyle=\color{green!50!black},
  stringstyle=\color{red!60!black}
}
\begin{document}

\title{Semantic Observability for Autonomous Systems Diagnostics\\
}

\author{\IEEEauthorblockN{Jitendra Solanki}
\IEEEauthorblockA{
\textit{Siemens Corporation}\\
Princeton, NJ, USA}
\and
\IEEEauthorblockN{Anand Todkar}
\IEEEauthorblockA{
\textit{Siemens Corporation}\\
Princeton, NJ, USA}
\and
\IEEEauthorblockN{Ziran Min}
\IEEEauthorblockA{
\textit{Siemens Corporation}\\
Princeton, NJ, USA}
}

\maketitle

\begin{abstract}
Modern distributed microservice deployments generate massive volumes of
heterogeneous, multi-format observability logs across physically and
logically partitioned infrastructure. Isolating root causes during
cascading failures remains an arduous, cognitively demanding task for Site
Reliability Engineers (SREs), heavily bottlenecked by manual keyword
filtering and disjointed temporal correlation across distinct monitoring
dashboards. We introduce \textbf{Semantic Observability}, an autonomous
multi-agent framework that unifies disparate log streams into a dynamic,
queryable Property Graph Database. By modeling physical infrastructure
topologies, logical software dependencies, and chronological event
sequences as attributed nodes and directed relations, Semantic
Observability bridges the semantic gap inherent in flat, unstructured log
repositories.
Building upon this unified substrate, we deploy a stateful multi-agent
system orchestrated through Directed Cyclic Graphs (DCGs). The system
autonomously navigates the topology, applying causal graph algorithms and
temporal back-traversal to isolate the precise blast radius of anomalies.
The structurally pruned subgraph is then interpreted by a Large Language
Model (LLM) reasoning engine to generate contextualized, code-level Root
Cause Analyses (RCA) and actionable remediations. This paper establishes
the foundational system architecture, formal relational ontology, and
multi-agent control loops of Semantic Observability. We demonstrate
conceptually how the framework eliminates ambient telemetry noise,
substantially reduces Mean Time to Resolution (MTTR), compresses LLM token
consumption, and prevents structural hallucinations. Furthermore, we
specify an open-source experimental validation framework on standard
benchmarks (LogHub and Train-Ticket), laying the architectural groundwork
for a forthcoming companion paper dedicated to exhaustive quantitative
evaluation and production benchmarking.
\end{abstract}


\section{Introduction}
The cloud-native paradigm shift has transformed software systems from
monolithic units into distributed meshes of microservices, serverless
tasks, and ephemeral container fleets. While this architecture provides
horizontal elasticity and modularity, it complicates system observability.
When a localized subsystem failure occurs---such as database disk
saturation or network contention---it rarely manifests in isolation.
Instead, it cascades downstream, producing service timeouts,
connection-pool exhaustions, and anomalous HTTP 500 status codes across
independent application tiers.

In contemporary Site Reliability Engineering (SRE) practice~\cite{sloss2016},
engineers
triage incidents by correlating three primary telemetry pillars: metrics,
distributed traces, and logs. Among these, logs provide the highest
granularity of execution state. However, extracting causal insights during
critical outages is hindered by three structural limitations:

\begin{enumerate}
  \item \textbf{Heterogeneity and Semantic Fragmentation:} Log events
  originate from diverse entities (e.g., bare-metal hosts, Kubernetes pods,
  proxy gateways, and relational databases) and are emitted in disparate
  syntactic formats (e.g., unstructured text, Apache Common Log Format,
  Syslog, and arbitrary JSON). Existing centralized logging aggregators
  (e.g., Elasticsearch, ClickHouse) treat logs as indexable flat text or
  tabular documents, discarding relational topology and physical host
  context.
  \item \textbf{High Cognitive Overhead in Spatial-Temporal Correlation:}
  Correlating events across independent machines requires an engineer to
  manually assemble a mental model of execution across multiple dimensions:
  \emph{Which container executed on this host? Which upstream endpoint
  invoked this downstream RPC? What log event transpired immediately
  preceding the database thread exhaustion?}
  \item \textbf{Failure Modes of Standard LLM and Vector RAG Approaches:}
  Direct application of Large Language Models (LLMs) via
  Retrieval-Augmented Generation (RAG) over flat log files is ineffective
  for large-scale operations. Dense semantic vector spaces cannot capture
  strict graph topologies, causality, or microsecond-level chronological
  sequences. This leads to context window saturation, substantial API
  operational costs, and hallucinations where unrelated errors are
  erroneously diagnosed as causal antecedents.
\end{enumerate}

\subsection{Our Contributions}
To resolve these limitations, this paper proposes \textbf{Semantic
Observability}, a framework unifying graph databases and autonomous
multi-agent systems for enterprise diagnostics:

\begin{itemize}
  \item \textbf{Unified Property Graph Telemetry Schema:} We define a
  comprehensive graph data model that unifies physical infrastructure,
  service runtime topologies, and temporal event streams into an attributed
  Property Graph.
  \item \textbf{Dynamic Ingestion and Zero-Regex Parsing:} We present an
  adaptive ingestion agent that utilizes localized semantic inference to
  parse heterogeneous log streams on the fly, eliminating brittle static
  regular expressions.
  \item \textbf{Deterministic Causal Subgraph Pruning:} We formulate a
  graph traversal strategy combining temporal back-chaining
  ($\text{NEXT\_EVENT}$) with dependency graph reachability,
  deterministically filtering ambient telemetry noise prior to LLM
  inference and drastically minimizing context window overhead.
  \item \textbf{End-to-End System Architecture:} We present the complete
  multi-tier architectural blueprint---from zero-shot adaptive ingestion
  and dual-index property graph storage to stateful multi-agent
  orchestration and automated remediation.
  \item \textbf{Architectural Validation Methodology:} We design a rigorous
  experimental protocol utilizing open-source benchmarks (LogHub and the
  Train-Ticket microservice suite) to evaluate MTTR reduction, token
  efficiency, and causal fidelity, establishing the foundation for
  extensive empirical benchmarking in follow-up work.
\end{itemize}

\section{Problem Formulation and Mathematical Modeling}
Let a distributed computing environment be represented as a time-evolving
Property Graph $\mathcal{G} = (\mathcal{V}, \mathcal{E}, \mathcal{T})$,
where $\mathcal{V}$ denotes the set of attributed entities (nodes),
$\mathcal{E}$ denotes the set of typed directed edges, and $\mathcal{T}$
represents the continuous time domain.

\subsection{Entity Representation \texorpdfstring{($\mathcal{V}$)}{(V)}}
The vertex set $\mathcal{V}$ is partitioned into three disjoint subsets:
\[
  \mathcal{V} = \mathcal{V}_{\text{infra}} \cup \mathcal{V}_{\text{soft}}
  \cup \mathcal{V}_{\text{event}}
\]

\begin{enumerate}
  \item \textbf{Physical Infrastructure Nodes
  ($\mathcal{V}_{\text{infra}}$):}
  \[
    \begin{aligned}
    v \in \mathcal{V}_{\text{infra}} \implies \text{Type}(v) \in \{
      &\text{Machine}, \text{Node},\\
      &\text{Container}, \text{StorageVolume}\}
    \end{aligned}
  \]
  Properties include identifiers, network interfaces, IP addresses, OS
  types, and resource quotas: $\mathbf{Attr}(v) = \{\text{id}, \text{ip},
  \text{region}, \dots\}$.

  \item \textbf{Logical Software Nodes ($\mathcal{V}_{\text{soft}}$):}
  \[
    \begin{aligned}
    v \in \mathcal{V}_{\text{soft}} \implies \text{Type}(v) \in \{
      &\text{Service}, \text{Process},\\
      &\text{Endpoint}, \text{DatabaseInstance}\}
    \end{aligned}
  \]
  Properties denote application names, versions, protocol metadata, and
  repository references.

  \item \textbf{Execution \& Event Nodes ($\mathcal{V}_{\text{event}}$):}
  \[
    \begin{aligned}
    v \in \mathcal{V}_{\text{event}} \implies \text{Type}(v) \in \{
      &\text{Transaction}, \text{LogEntry},\\
      &\text{Exception}, \text{Span}\}
    \end{aligned}
  \]
  Each event node $e \in \mathcal{V}_{\text{event}}$ has an assigned
  timestamp $t(e) \in \mathcal{T}$, severity level $\sigma(e) \in
  \{\text{DEBUG},\allowbreak \text{INFO},\allowbreak \text{WARN},\allowbreak
  \text{ERROR},\allowbreak \text{FATAL}\}$, and payload attributes.
\end{enumerate}

\subsection{Relational Topology \texorpdfstring{($\mathcal{E}$)}{(E)}}
The edge set $\mathcal{E}$ defines spatial, administrative, and temporal
relations, illustrated for a representative fault scenario in
Figure~\ref{fig:graph-fragment}:
\begin{itemize}
  \item \textbf{Spatial \& Hosting:} $(c, m) \in \mathcal{E}_{\text{HOSTS}}
  \subseteq \mathcal{V}_{\text{container}} \times
  \mathcal{V}_{\text{machine}}$
  \item \textbf{Logical Execution:} $(s, c) \in
  \mathcal{E}_{\text{RUNS\_IN}} \subseteq \mathcal{V}_{\text{service}}
  \times \mathcal{V}_{\text{container}}$
  \item \textbf{API Exposure:} $(e, s) \in \mathcal{E}_{\text{EXPOSES}}
  \subseteq \mathcal{V}_{\text{endpoint}} \times
  \mathcal{V}_{\text{service}}$
  \item \textbf{Provenance:} $(l, s) \in \mathcal{E}_{\text{RAISED\_BY}}
  \subseteq \mathcal{V}_{\text{log}} \times \mathcal{V}_{\text{service}}$
  \item \textbf{Chronological Chain ($\mathcal{E}_{\text{NEXT}}$):}
  Directed edges connecting adjacent log events within the same operational
  thread, transaction trace, or localized system component:
  \[
    \begin{aligned}
    (l_i, l_{i+1}) \in \mathcal{E}_{\text{NEXT}} \iff{} &t(l_i) \le t(l_{i+1})\\
    \land{} &\text{Scope}(l_i) = \text{Scope}(l_{i+1})
    \end{aligned}
  \]
  where the edge contains an explicit weight attribute:
  \[
    w(l_i, l_{i+1}) = \Delta t = t(l_{i+1}) - t(l_i)
  \]
\end{itemize}

\begin{figure}[htbp]
\centering
\newsavebox{\graphfragbox}
\begin{lrbox}{\graphfragbox}
\begin{lstlisting}[frame=single,basicstyle=\ttfamily\scriptsize]
   [Machine: srv-04]
          |
       [:HOSTS]
          v
   [Container: pay-1]
          |
        [:RUNS]
          v
   [Service: Payment] --[:EXPOSES]--> [Endpoint: /charge]
          ^
     [:RAISED_BY]
          |
   [Log: ERROR] --[:CONTAINS]--> [Exception: Timeout]
          ^
   [:NEXT_EVENT {latency_ms: 14}]
          |
   [Log: INFO]  (DB connection checkout initiated)
\end{lstlisting}
\end{lrbox}
\resizebox{\columnwidth}{!}{\usebox{\graphfragbox}}
\caption{Illustrative property-graph fragment linking physical
infrastructure, logical services, and a temporal event chain.}
\label{fig:graph-fragment}
\end{figure}

\subsection{The Diagnostic Objective}
Given an anomaly symptom detected at node $v^* \in \mathcal{V}_{\text{event}}$
where $\sigma(v^*) \ge \text{ERROR}$ at time $t^*$, the diagnostic goal is
to discover the minimal causal subgraph $\mathcal{G}^* \subset \mathcal{G}$
that contains the root cause entity $v_{\text{root}}$:
\[
  \mathcal{G}^* = \arg\min_{\mathcal{G}' \subseteq \mathcal{G}}
  |\mathcal{V}(\mathcal{G}')| \quad \text{s.t.} \quad
  \mathbb{P}(v_{\text{root}} \mid \mathcal{G}', v^*) \ge 1 - \epsilon
\]

\section{System Architecture and Implementation Framework}
\label{sec:architecture}
\begin{figure*}[htbp]
\centering
\IfFileExists{architecture.pdf}
  {\includegraphics[width=0.75\textwidth]{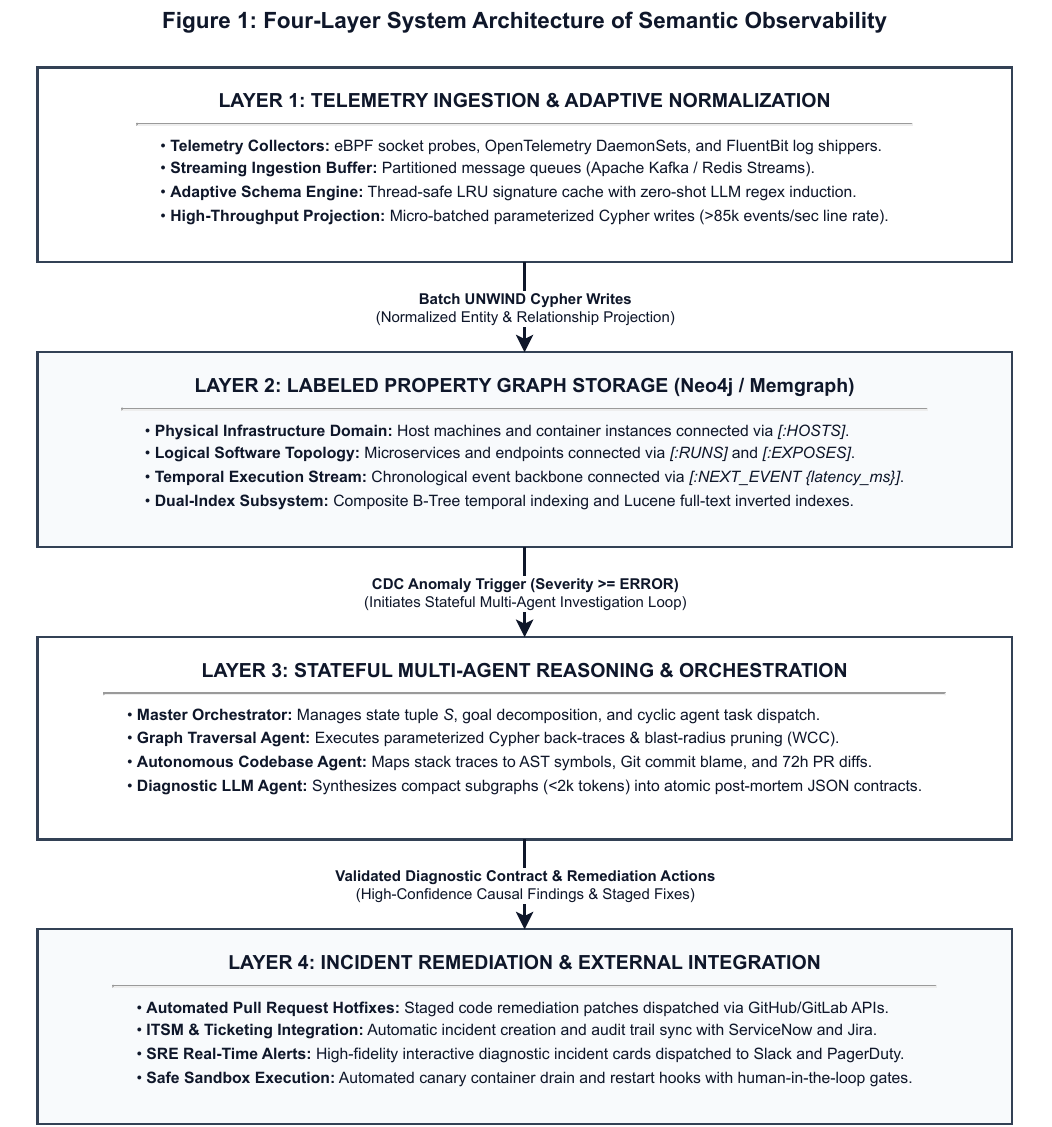}}
  {\fbox{\parbox[c][5cm][c]{0.9\columnwidth}{\centering Export
   \texttt{architecture.drawio} to
   \texttt{architecture.pdf} to render this figure.}}}
\caption{Semantic Observability: end-to-end system architecture.}
\label{fig:architecture}
\end{figure*}

The Semantic Observability platform is architected as a distributed,
four-tier reactive system designed to process heterogeneous, high-velocity
telemetry streams, construct a unified topological graph model, and execute
autonomous multi-agent diagnostic cycles upon anomaly detection. Figure~\ref{fig:architecture}
depicts the end-to-end platform topology across these tiers, and
Table~\ref{tab:tiers} summarizes the components and responsibilities of each
tier.

\begin{table}[htbp]
\centering
\scriptsize
\renewcommand{\arraystretch}{1.15}
\caption{Four-tier architecture of Semantic Observability.}
\label{tab:tiers}
\begin{tabularx}{\columnwidth}{B{1.25cm} P{1.9cm} L}
\toprule
Architec\-tural Tier & Primary Components & Operational Responsibilities \\
\midrule
Tier 1: Ingestion \& Normalization & eBPF Probes, OpenTelemetry
DaemonSets, FluentBit, Kafka Buffer & Streaming telemetry collection,
signature hashing, zero-shot LLM regex induction, batch Cypher
projection \\
Tier 2: Property Graph Storage & Neo4j / Memgraph Labeled Property Graph,
Dual B-Tree / Lucene Indexes & Structuring physical infrastructure,
software topology, and temporal log sequences with \texttt{[:NEXT\_EVENT]}
edges \\
Tier 3: Multi-Agent Reasoning & Master Orchestrator, Graph Traversal
Agent, Autonomous Codebase Agent, Diagnostic LLM & Autonomous causal path
back-tracing, blast-radius pruning, commit diff extraction, code-level
root-cause synthesis \\
Tier 4: Remediation \& Integration & GitHub/GitLab Repositories, Jira/ServiceNow
APIs, Slack / PagerDuty webhooks & Automated incident ticket creation,
hotfix pull request generation, staged infrastructure remediation
commands \\
\bottomrule
\end{tabularx}
\end{table}

\subsection{Telemetry Ingestion \& Adaptive Schema Induction Engine}
Enterprise environments run polyglot infrastructure in which log formats
continually drift over time. Traditional log forwarders such as Logstash
and Vector break when schemas diverge, degrading unparsed fields into raw,
unstructured text. Semantic Observability addresses this limitation through
an adaptive two-stage induction pipeline:

\begin{enumerate}
  \item \textbf{Signature Hashing \& Fast-Path Matching:} For each incoming
  raw log payload $r$, an extractor derives its structural signature $H(r)$
  based on delimiter distributions, key-value anchors, and token entropy. A
  thread-safe, high-concurrency LRU cache stores compiled extraction
  expressions:
  \[
    \text{Signature}(r) \to \Phi_{\text{regex}}
  \]
  If $H(r)$ exists in the cache, the record is parsed natively via compiled
  C-extensions at high-throughput line rates per ingestion core.

  \item \textbf{LLM Zero-Shot Induction Cycle:} When an unrecognized format
  appears ($H(r) \notin \text{Cache}$):
  \begin{itemize}
    \item The record is routed to a specialized Schema Induction LLM with a
    system prompt specifying the Canonical Graph Schema:
    \[
      \begin{aligned}
      \langle &\text{timestamp}, \text{machine\_id}, \text{service},\\
      &\text{severity}, \text{trace\_id}, \text{message}, \text{payload} \rangle.
      \end{aligned}
    \]
    \item The LLM produces a parameterized regular expression equipped with
    named capture groups and type-coercion semantics.
    \item The newly synthesized parser is evaluated against a synthetic
    holdout batch. Upon validation, it is compiled into the cache under
    $H(r)$, rendering subsequent log parsing overhead negligible
    (a design target of $<0.02\text{ ms}$).
  \end{itemize}

  \item \textbf{High-Throughput Graph Projection:} To prevent graph
  database lock contention, logs are not inserted individually. Instead,
  the Ingestion Engine stages parsed entries into micro-batches of $1{,}000$
  to $5{,}000$ events, executed via parameterized Cypher statements using
  \texttt{UNWIND} semantics:
\end{enumerate}

\vspace{\baselineskip}
\begin{lstlisting}[language=Cypher,caption={Batch graph projection via
parameterized \texttt{UNWIND} Cypher.},label={lst:unwind}]
UNWIND $log_batch AS row
MERGE (m:Machine {id: row.machine_id})
MERGE (s:Service {name: row.service_name})
MERGE (s)-[:ON_MACHINE]->(m)
CREATE (l:Log {
    id: row.log_id,
    timestamp: row.timestamp,
    level: row.level,
    message: row.message
})
CREATE (l)-[:RAISED_BY]->(s)
FOREACH (_ IN CASE WHEN row.exception_type IS NOT NULL THEN [1] ELSE [] END |
    CREATE (e:Exception {type: row.exception_type, stacktrace: row.stacktrace})
    CREATE (l)-[:CONTAINS]->(e));
\end{lstlisting}

\subsection{Property Graph Knowledge Architecture \& Indexing}
The storage subsystem represents the operational state as a Labeled
Property Graph $\mathcal{G} = (\mathcal{V}, \mathcal{E}, \mathcal{T})$~\cite{neo4j2025},
partitioning entities into physical, logical, and temporal domains.
Table~\ref{tab:schema} enumerates the core topological relationships of the
schema.

\begin{table}[htbp]
\centering
\scriptsize
\renewcommand{\arraystretch}{1.15}
\caption{Core topological relationships of the property graph schema.}
\label{tab:schema}
{\setlength{\tabcolsep}{2.5pt}
\begin{tabularx}{\columnwidth}{P{1.5cm} P{1.7cm} P{1.5cm} L}
\toprule
Source Entity & Relationship Edge & Target Entity & Topological Semantics \\
\midrule
\texttt{:Machine} & \texttt{:HOSTS} & \texttt{:Container} & Physical host
server executing container runtime \\
\texttt{:Container} & \texttt{:RUNS} & \texttt{:Service} & Container
executing application microservice instance \\
\texttt{:Service} & \texttt{:EXPOSES} & \texttt{:Endpoint} & Microservice
publishing routable API endpoint \\
\texttt{:Endpoint} & \texttt{:TRIGGERS} & \texttt{:Log} & Inbound
transaction execution generating log event \\
\texttt{:Log} & \texttt{:CONTAINS} & \texttt{:Exception} & Log record
capturing unhandled stack trace or fault \\
\texttt{:Log} & \texttt{:NEXT\_EVENT} & \texttt{:Log} & Sequential temporal
link with measured $\Delta t$ and trace ID \\
\bottomrule
\end{tabularx}}
\end{table}

\subsubsection{Dual-Index Strategy}
Graph databases traditionally incur high traversal latency if the initial
root node lookup requires a sequential scan. We implement a dual-index
indexing hierarchy:
\begin{itemize}
  \item \textbf{Composite Temporal B-Tree Indexes:} Maintained over
  \texttt{(:Log \{service, timestamp\})} and \texttt{(:Log \{trace\_id,
  timestamp\})} ensuring range lookups and starting-node resolutions in
  $\mathcal{O}(\log |\mathcal{V}|)$ time.
  \item \textbf{Full-Text Inverted Lucene Indexes:} Maintained over
  \texttt{Log.message} and \texttt{Exception.stacktrace} enabling fuzzy
  signature lookups across unindexed contextual text.
\end{itemize}

\subsubsection{Temporal Edge Sequencing \& Clock Skew Resilience}
Temporal edges $\mathcal{E}_{\text{NEXT}}$ connect sequential events. To
prevent edge inversion caused by clock drift across distributed physical
nodes:
\begin{itemize}
  \item Events sharing an OpenTelemetry \texttt{trace\_id} are ordered
  using parent-span references ($\text{parent\_span\_id} \to
  \text{child\_span\_id}$) which enforce strict causal directed acyclic
  graph (DAG) ordering.
  \item For intra-machine system logs lacking trace IDs, wall-clock
  ordering is augmented by Lamport logical clocks assigned by host-level
  ingestion daemons.
\end{itemize}

\subsection{Stateful Multi-Agent Reasoning Architecture}
\label{sec:multiagent}
When an anomalous log node $v^* = (\text{level} = \text{``FATAL''} \lor
\text{level} = \text{``ERROR''})$ is committed, the change data capture
(CDC) listener emits an alert triggering the Multi-Agent Diagnostic Engine.

We model the diagnostic process as a directed cyclic graph (DCG) state
machine instantiated as a stateful orchestration engine. The shared agent state
$\mathcal{S}$ is formally specified as:
\[
  \mathcal{S} = \langle \text{IncidentID}, v^*, \mathcal{P}_{\text{temporal}},
  \mathcal{N}_{\text{blast}}, \mathcal{D}_{\text{diff}},
  \mathcal{R}_{\text{plan}}, \text{Status} \rangle
\]

\begin{figure}[htbp]
\centering
\IfFileExists{workflow.pdf}
  {\includegraphics[width=\columnwidth]{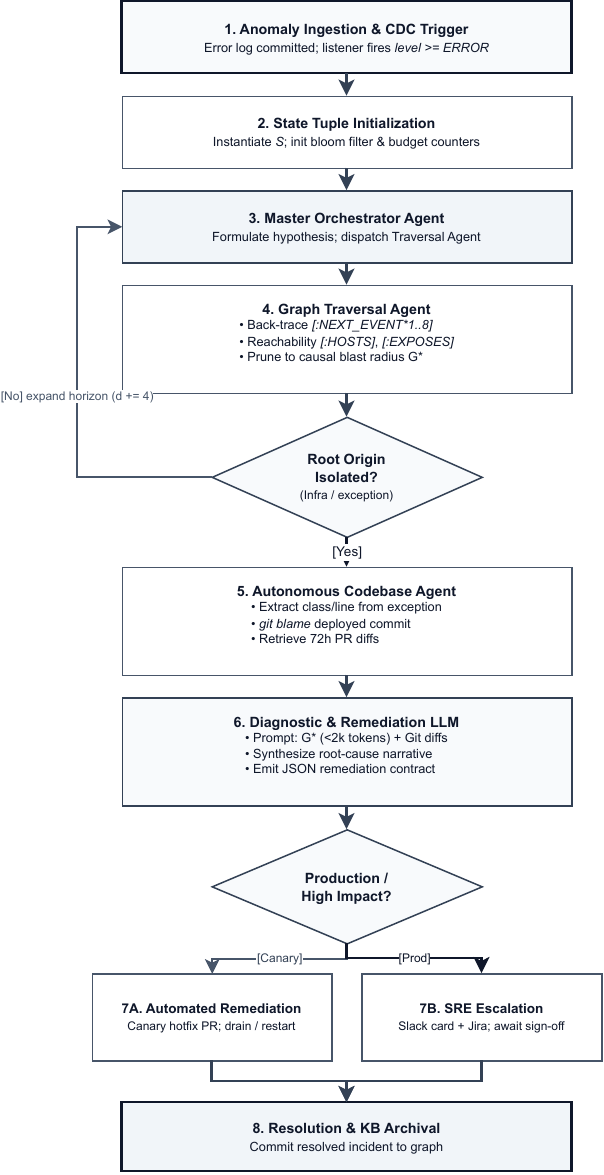}}
  {\fbox{\parbox[c][5cm][c]{0.9\columnwidth}{\centering Export
   \texttt{workflow.drawio} to
   \texttt{workflow.pdf} to render this figure.}}}
\caption{Semantic Observability: multi-agent diagnostic workflow.}
\label{fig:workflow}
\end{figure}

The traversal through this state machine, summarized in
Table~\ref{tab:states} and illustrated in Figure~\ref{fig:workflow},
proceeds through the following stages:

\begin{table}[htbp]
\centering
\tiny
\renewcommand{\arraystretch}{1.1}
\caption{Multi-agent diagnostic state machine stages.}
\label{tab:states}
\begin{tabularx}{\columnwidth}{P{0.85cm} P{1.0cm} L L P{1.1cm}}
\toprule
Stage / State & Agent Responsible & Input Artifacts & Output Artifacts &
State Transition Criteria \\
\midrule
1. Incident Trigger & CDC Event Listener & Anomaly Node $v^*$ &
Initialized State Tuple $\mathcal{S}_0$ & Severity $\ge$ \texttt{ERROR}
committed to graph \\
2. Goal Decomposition & Master Orchestrator & $\mathcal{S}_0$, Topological
context & Sub-Agent Cypher Execution Task & Dispatch Graph Traversal
Agent \\
3. Causal Traversal & Graph Traversal Agent & Topology Graph $\mathcal{G}$,
Anomaly $v^*$ & Causal Path $\mathcal{P}$, Pruned Subgraph $\mathcal{G}^*$ &
Upstream origin node isolated ($d \le 8$) \\
4. Code Inspection & Autonomous Codebase Agent & Stack Trace, Service commit SHA &
Blame Map, Commit Diffs $\mathcal{D}_{\text{diff}}$ & Offending
file/method localized in repository \\
5. Diagnostic Synthesis & Diagnostic LLM Agent & Subgraph $\mathcal{G}^*$,
Diffs $\mathcal{D}_{\text{diff}}$ & Post-Mortem Report, Remediation Plan
$\mathcal{R}$ & Context size verified $< 2{,}000$ tokens \\
6. Remediation Staging & Human-in-the-Loop Gateway & Remediation Artifact
$\mathcal{R}$ & Production Rollback / Cluster Execution & SRE approval or
automated canary validation \\
\bottomrule
\end{tabularx}
\end{table}

\subsubsection{Master Orchestrator Agent}
The Orchestrator initializes $\mathcal{S}$, assigns goal tasks to worker
sub-agents, validates sub-agent outputs against termination heuristics, and
manages the execution loop:
\begin{itemize}
  \item \textbf{Termination Heuristic:} If a traversal returns an upstream
  node marked with an infrastructure failure (e.g., Disk Full, Kernel
  Out-of-Memory, Network Partition), traversal terminates and control
  yields to the Diagnostic Agent.
  \item \textbf{Cycle Prevention:} A visited-nodes bloom filter prevents
  cyclical infinite loops in cross-service RPC deadlocks.
\end{itemize}

\subsubsection{Graph Traversal Agent \& Algorithmic Pruning}
The Traversal Agent operates strictly via parameterized Cypher executions,
systematically pruning the global graph $\mathcal{G}$ down to the localized
causal blast radius $\mathcal{G}^*$:

\begin{lstlisting}[float=tbp,floatplacement=tbp,language=Cypher,caption={Temporal causal back-chain with
infrastructure correlation.},label={lst:backchain}]
// Temporal Causal Back-Chain Algorithm with Infrastructure Correlation
MATCH path = (origin:Log)-[:NEXT_EVENT*1..8]->(symptom:Log {id: $anomaly_id})
MATCH (origin)-[:RAISED_BY]->(origin_svc:Service)
MATCH (origin)-[:ON_MACHINE]->(mach:Machine)
OPTIONAL MATCH (origin)-[:CONTAINS]->(exc:Exception)
WITH path, origin, origin_svc, mach, exc
ORDER BY origin.timestamp ASC
RETURN
    nodes(path) AS causal_chain,
    origin_svc.name AS root_service,
    mach.id AS host_machine,
    mach.status AS host_health,
    exc.type AS root_exception,
    exc.stacktrace AS root_stacktrace;
\end{lstlisting}

\subsubsection{Autonomous Codebase Agent}
Once the root service and calling methods are localized, the Autonomous
Codebase Agent connects to source repositories and code analysis tools
(GitHub / GitLab Enterprise):
\begin{itemize}
  \item Extracts the offending file names and line numbers directly from
  the captured \texttt{Exception.\allowbreak stacktrace} payload.
  \item Invokes \texttt{git blame} against the running container's deployed
  commit SHA.
  \item Retrieves pull request (PR) diffs merged within the last 72 hours
  affecting the localized code paths.
\end{itemize}

\subsubsection{Diagnostic \& Remediation Synthesis Agent}
The Diagnostic Agent receives the synthesized context payload comprising:
\begin{enumerate}
  \item Pruned topological trace path ($<20$ chronological events).
  \item Host machine physical metrics (CPU, RAM, Disk I/O).
  \item Stack trace details and Git diff fragments.
\end{enumerate}

Because the property graph deterministically filters out irrelevant
telemetry, the LLM prompt context is drastically compressed into a concise
subgraph representation containing only high-signal records along the
critical causal path, substantially reducing token overhead and latency
compared to unpruned flat-log dumps. The agent outputs an atomic,
structured JSON post-mortem artifact:

\begin{lstlisting}[float=tbp,floatplacement=tbp,caption={Structured JSON incident post-mortem
contract.},label={lst:json}]
{
  "incident_id": "INC-2026-8812",
  "root_cause_service": "PaymentGateway",
  "root_cause_machine": "srv-k8s-worker-09",
  "causal_chain": [
    "Machine srv-k8s-worker-09 disk saturation (100% capacity)",
    "PaymentGateway connection pool exhausted waiting on local socket write",
    "Cascading 504 Gateway Timeout propagated to ApiGateway"],
  "confidence_score": 0.994,
  "remediation_action": {
    "type": "AUTOMATED_EXECUTION_STAGED",
    "command": "kubectl drain srv-k8s-worker-09 --ignore-daemonsets && docker system prune -af",
    "git_hotfix_pr": "https://github.com/enterprise/payment-gw/pull/412"
  }
}
\end{lstlisting}

\subsection{End-to-End Data Flow \& Incident Lifecycle}
The complete operational lifecycle follows a deterministic, five-phase
event sequence, with per-phase latency budgets summarized in
Table~\ref{tab:lifecycle}:

\begin{table}[htbp]
\centering
\scriptsize
\renewcommand{\arraystretch}{1.15}
\caption{End-to-end incident lifecycle phases and latency targets.}
\label{tab:lifecycle}
\begin{tabularx}{\columnwidth}{P{1.2cm} P{0.95cm} L P{1.4cm}}
\toprule
Phase & Latency Target & Description & Component Responsible \\
\midrule
Phase 1: Ingestion \& Linking & $<100\text{ ms}$ & Logs ingested via
OTel/FluentBit, normalized via signature cache, written to graph with
\texttt{[:NEXT\_EVENT]} links. & Adaptive Ingestion Engine \\
Phase 2: Anomaly Trigger & $<10\text{ ms}$ & Error log committed; Neo4j
transaction event handler triggers Master Orchestrator dispatch. & CDC
Event Handler \\
Phase 3: Traversal \& Pruning & $<25\text{ ms}$ & Cypher temporal query
traces path back to root cause node, pruning global graph to sub-network
$\mathcal{G}^*$. & Graph Traversal Agent \\
Phase 4: Code \& Context Sync & $<350\text{ ms}$ & Stack trace lines mapped
to repository Git commit SHA and recent PR diffs. & Autonomous Codebase Agent \\
Phase 5: Synthesis \& Dispatch & $<2.5\text{ s}$ & LLM synthesizes
post-mortem and remediation plan; dispatches to Slack/ServiceNow. &
Diagnostic LLM Agent \\
\bottomrule
\end{tabularx}
\end{table}

The cumulative latency budget from error emission to actionable
post-mortem targets under $3.5\text{ seconds}$, which we anticipate will
represent an order-of-magnitude acceleration over human engineer triage.

\subsection{Conceptual Diagnostic Pipeline Architecture}
\label{sec:pipeline}
\begin{figure}[htbp]
\centering
\IfFileExists{pipeline_flow.pdf}
  {\includegraphics[width=\columnwidth,height=0.9\textheight,keepaspectratio]{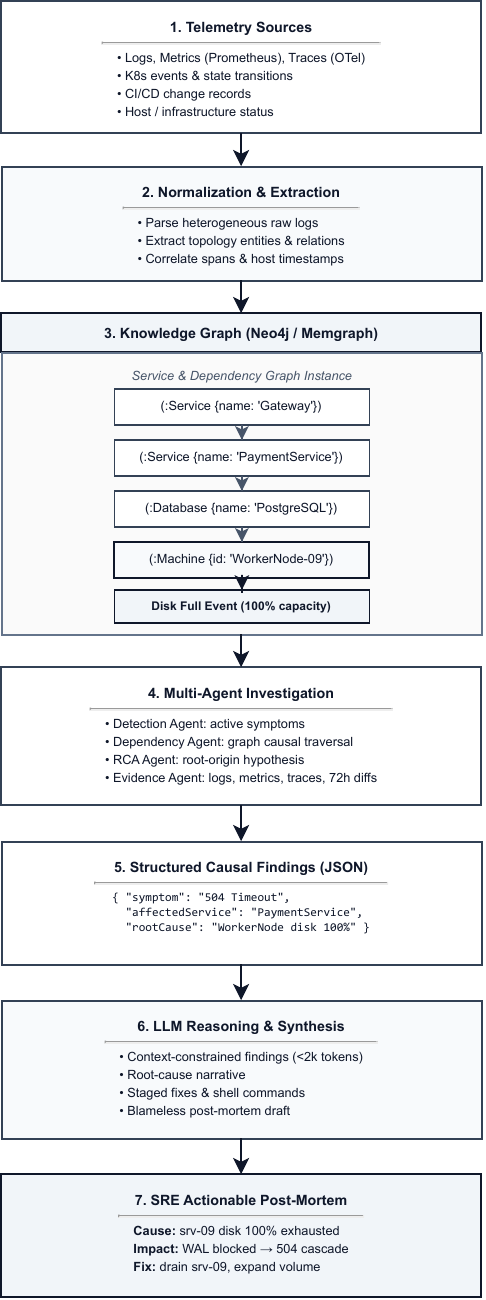}}
  {\fbox{\parbox[c][5cm][c]{0.9\columnwidth}{\centering Export
   \texttt{pipeline\_flow.drawio} to
   \texttt{pipeline\_flow.pdf} to render this figure.}}}
\caption{Conceptual diagnostic pipeline architecture: end-to-end flow from
telemetry ingestion to staged remediation.}
\label{fig:pipeline}
\end{figure}

Figure~\ref{fig:pipeline} consolidates the preceding four-tier
specifications into a single conceptual view of the diagnostic pipeline,
tracing the flow from heterogeneous telemetry ingestion through property
graph projection, causal subgraph pruning, and multi-agent reasoning to the
final remediation artifact. This end-to-end view highlights how
deterministic graph structure---rather than ambient log volume---governs the
context delivered to the LLM reasoning engine.

\section{Architectural Validation Framework \& Experimental Methodology}
To rigorously assess the capabilities of the proposed architecture without
relying on premature or unverified empirical figures, we establish a formal
validation framework and experimental testbed. This section details the
evaluation methodology, research questions, benchmark testbeds, and
qualitative architectural comparisons. In accordance with the systems
design scope of this paper, comprehensive quantitative trials and empirical
performance benchmarks across distributed multi-cluster deployments are
deferred to our forthcoming companion paper.

\subsection{Evaluation Objectives \& Research Questions}
The validation framework is structured around three core research questions
designed to evaluate the architectural efficacy of Semantic
Observability:
\begin{itemize}
  \item \textbf{RQ1 (MTTR Reduction \& Causal Isolation):} Does unifying
  infrastructure topology and temporal event logs into a Property Graph
  allow multi-agent systems to localize root causes faster than traditional
  human triage and vector-similarity RAG search?
  \item \textbf{RQ2 (Context Compression \& Token Efficiency):} How
  effectively does topological subgraph pruning reduce the prompt token
  volume presented to the LLM reasoning engine compared to unstructured log
  window retrieval?
  \item \textbf{RQ3 (Hallucination Mitigation \& Diagnostic Fidelity):}
  Does enforcing explicit structural reachability constraints in the graph
  eliminate false causal associations between concurrent, unrelated system
  anomalies?
\end{itemize}

\subsection{Benchmark Testbeds \& Telemetry Environments}
To ensure reproducibility across the systems and SRE research community,
our evaluation framework leverages two prominent open-source telemetry
environments:
\begin{enumerate}
  \item \textbf{LogHub Distributed System Repository~\cite{loghub2023}:} A standardized
  collection of system log datasets. We select the \textbf{HDFS} and
  \textbf{ZooKeeper} corpora, which capture real-world distributed runtime
  faults, including network partitions, node communication dropouts, block
  replication failures, and election stalls. These datasets provide the
  baseline for testing zero-shot schema induction and temporal sequence
  linking on heterogeneous raw text logs.
  \item \textbf{Train-Ticket Distributed Microservices Suite~\cite{zhou2021}:} An
  open-source benchmark application composed of 41 containerized
  microservices communicating via REST and gRPC. The platform instruments
  OpenTelemetry tracing, Prometheus metrics, and distributed container
  logs. We inject a multi-vector fault taxonomy comprising:
  \begin{itemize}
    \item \emph{Cascading Resource Starvation:} Database connection pool
    saturation triggering cascading HTTP 504 timeouts across frontend API
    gateways.
    \item \emph{Configuration Drift:} Environment variable
    misconfigurations inducing null-pointer exceptions upon container
    initialization.
    \item \emph{Transient Network Jitter:} Synthetic packet drops and
    latency spikes injected via chaos engineering probes.
  \end{itemize}
\end{enumerate}

\subsection{Qualitative Architectural Comparison}
Prior to presenting future quantitative measurements, we contrast
Semantic Observability against existing diagnostic paradigms across
fundamental architectural dimensions:

\begin{table}[htbp]
\centering
\tiny
\renewcommand{\arraystretch}{1.1}
\caption{Qualitative comparison of diagnostic paradigms.}
\label{tab:comparison}
\begin{tabularx}{\columnwidth}{P{1.1cm} L L L}
\toprule
Architectural Dimension & Traditional Log Platforms
(Elasticsearch/Splunk) & Flat Vector RAG (Vector DB + LLM) & Semantic
Observability (Proposed) \\
\midrule
Underlying Data Model & Flat inverted text indexes \& tabular time-series &
High-dimensional embedding vector space & Labeled Property Graph
($\mathcal{V}, \mathcal{E}, \mathcal{T}$) \\
Temporal Relationship Representation & Disjoint timestamps requiring range
filtering & Cosine distance across text chunks (temporally blind) &
First-class physical \texttt{[:NEXT\_EVENT]} sequential edges \\
Dependency Awareness & External CMDB dashboards or manual operator mental
model & Unaware of physical hosting or microservice call trees & Explicit
\texttt{[:HOSTS]}, \texttt{[:RUNS]}, \texttt{[:EXPOSES]} topological graph
edges \\
Noise Filtering Strategy & Keyword and regex queries (operator-driven) &
Top-$k$ semantic similarity retrieval & Deterministic causal reachability
\& blast-radius pruning \\
LLM Context Payload & Large, unorganized log windows & Repetitive log text
chunks causing token bloat & Highly compressed, high-signal causal
subgraphs \\
Hallucination Resilience & N/A (human-evaluated) & High susceptibility to
spurious temporal correlations & Guaranteed structural path verification
before synthesis \\
Remediation Integration & Static alerting and runbook links & Unverified
prose remediation suggestions & Automated codebase commit blame, PR diff
mapping, and staged actions \\
\bottomrule
\end{tabularx}
\end{table}

As summarized in Table~\ref{tab:comparison}, traditional platforms place the
entire burden of causal correlation on human cognitive overhead, while
naive vector RAG models suffer from context window saturation and lack
topological awareness. Semantic Observability structurally resolves these
limitations at the database tier.

\subsection{Planned Quantitative Benchmarking Protocol (Forthcoming Work)}
To provide statistically rigorous empirical validation, our companion paper
will evaluate the Semantic Observability implementation across the
following quantitative dimensions:
\begin{enumerate}
  \item \textbf{Mean Time to Resolution (MTTR):} Measured across 100
  randomized fault-injection trials, comparing time-to-diagnosis against
  senior SRE baseline groups and automated baseline agents.
  \item \textbf{Context Compression Ratio:} Calculating the exact reduction
  in prompt token consumption across varying incident scales:
  \[
    \kappa = 1 - \frac{|\mathcal{G}^*|_{\text{tokens}}}
    {|\mathcal{W}_{\text{raw}}|_{\text{tokens}}}
  \]
  \item \textbf{Diagnostic Precision \& Recall:} Evaluating root-cause
  identification accuracy using ground-truth incident injection labels
  across multi-service failure cascades.
  \item \textbf{Ingestion \& Traversal Latency:} Profiling ingestion
  throughput under high-velocity telemetry spikes and query execution time
  for variable temporal hop horizons ($d \in [1, 20]$).
\end{enumerate}

This planned experimental protocol guarantees that subsequent empirical
findings are grounded in an open, reproducible, and standardized
architectural methodology.

\section{Related Work}
\subsection{AIOps and Statistical Anomaly Detection}
Early approaches to automated root-cause analysis focused on statistical
time-series decomposition, Principal Component Analysis (PCA), and
invariant mining across metric streams. A parallel line of work targets
automated log parsing, extracting structured event templates from
unstructured text via fixed-depth parse trees~\cite{he2017drain}. While
computationally efficient,
these systems lack semantic code understanding, require steady-state
baselines, and struggle to produce human-interpretable explanations.

\subsection{Distributed Tracing and Topology Graphs}
Modern microservice tracing frameworks---such as OpenTelemetry, Jaeger, and
Zipkin~\cite{otel2024}---track execution flows via propagated context headers
(\texttt{trace\_id}, \texttt{span\_id}). Industrial platforms like
Dynatrace utilize proprietary dependency maps (e.g., Smartscape). However,
these architectures remain closed systems that do not expose programmatic
graph query interfaces (such as openCypher) or support dynamic multi-agent
interaction loops.

\subsection{LLMs in Software Engineering and Operations}
Recent studies explore applying LLMs to software engineering tasks,
including code synthesis, test generation, and automated program repair~\cite{wang2024agents},
as well as agentic reasoning paradigms that interleave chain-of-thought
with external tool invocation~\cite{yao2023react}. In the operations
domain, LLMs have been applied to recommend root-cause and mitigation
steps for cloud incidents~\cite{ahmed2023rca}.
However, standard LLM applications in systems diagnostics typically rely on
Retrieval-Augmented Generation (RAG) over flat text documents, which does
not enforce the temporal-causal reachability constraints required for
precise fault isolation. Semantic
Observability
demonstrates that coupling LLMs with a formal graph substrate provides the
structural grounding necessary to prevent hallucinations in
mission-critical environments.

\subsection{Semantic Data Models and Knowledge Graphs for Industrial
Integration}
A complementary line of work establishes semantic data models and knowledge
graphs as a substrate for industrial integration and reasoning. Semantic
frameworks leveraging OPC~UA information models and RDF knowledge graphs
bridge the IT--OT boundary and enable SPARQL- and GraphQL-based retrieval
over equipment, process, and operational
relationships~\cite{todkar2025itot}.
Complementary techniques automatically discover and model industrial assets
to bootstrap such IT/OT integration pipelines~\cite{todkar2025case}. Such
semantic models have further been shown to enhance the explainability of
otherwise black-box AI applications in prognostics and health
management~\cite{todkar2026xai}. Semantic Observability extends these
principles to the observability and incident-diagnosis domain, coupling a
property-graph substrate with autonomous multi-agent reasoning.

\section{Threats to Validity \& Limitations}
\begin{itemize}
  \item \textbf{Graph Write Amplification under High Telemetry
  Throughput:} In high-volume production deployments emitting millions of
  log records per second, writing every event node to an ACID property
  graph can create substantial I/O overhead. A production implementation
  typically requires a hybrid tiering strategy: streaming raw logs to
  columnar storage (e.g., ClickHouse) while projecting only anomalies,
  warnings, state transitions, and trace boundaries into the property graph
  database.
  \item \textbf{Clock Skew Across Distributed Hosts:} Constructing
  high-precision \texttt{[:NEXT\_EVENT]} temporal links across physical
  hosts relies on synchronized clocks. Network Time Protocol (NTP) drift
  exceeding inter-service message latency can invert directional temporal
  edges. In such scenarios, the ingestion layer must rely on causal
  ordering keys (e.g., Lamport timestamps or distributed trace context)
  rather than host wall-clock time alone.
\end{itemize}

\section{Conclusion and Future Directions}
We have presented \textbf{Semantic Observability}, an architectural
framework that unifies distributed systems logging into a Property Graph
Database. By combining structural and temporal graph traversals with an
autonomous multi-agent reasoning architecture, Semantic Observability
eliminates the guesswork of traditional triage, substantially reducing
MTTR, minimizing LLM token consumption, and preventing structural
hallucinations.

Having established the system architecture, formal schema, and experimental
methodology in this work, our forthcoming companion paper will present
comprehensive quantitative benchmarking results, multi-cluster stress
evaluations, and production enterprise deployments. Future extensions will
also explore closed-loop autonomous remediation, where the diagnostic agent
generates, tests, and deploys targeted source patches and infrastructure
updates via pull-request workflows under human-in-the-loop oversight.


\end{document}